\documentclass[sigconf]{acmart}
\AtBeginDocument{%
  }

\copyrightyear{2026}
\acmYear{2026}
\setcopyright{cc}
\setcctype{by}
\acmConference[CIKM '26]{Proceedings of the 35th ACM International Conference on Information and Knowledge Management}{November 07--11, 2026}{Rome, Italy}
\acmBooktitle{Proceedings of the 35th ACM International Conference on Information and Knowledge Management (CIKM '26), November 07--11, 2026, Rome, Italy}
\acmDOI{10.1145/3799682.3840276}
\acmISBN{979-8-4007-2539-5/2026/11}

\usepackage{xcolor}
\usepackage{balance}
\usepackage{mathtools}
\usepackage{url}
\usepackage{amsfonts}
\usepackage{algorithmic}
\usepackage{textcomp}
\usepackage{algorithm}
\usepackage{multirow}
\usepackage{color}
\usepackage{subcaption}
\usepackage{enumitem}

\usepackage{amsmath}
\usepackage{hyperref}

\newcommand{\method}{$\text{\textsf{PROBE}}^{\text{Web}}$}

\definecolor{LinkBlue}{RGB}{6,69,173}

\newcommand{\codeurl}{%
\textcolor{LinkBlue}{%
\href{https://github.com/mindslab-cau/probe-web-cikm26}
{https://github.com/mindslab-cau/probe-web-cikm26}}}

\newcommand{\videourl}{%
\textcolor{LinkBlue}{%
\href{https://tinyurl.com/PROBE-WEB-DEMO}
{https://tinyurl.com/PROBE-WEB-DEMO}}}

\begin{document}

\title{{\method}: An Interactive System for Probing Evaluation Landscapes of Knowledge Graph Completion Models}

\author{Sooho Moon}
\affiliation{%
  \institution{Chung-Ang University}
  \city{Seoul}
  \country{Korea}}
\email{moonwalk725@cau.ac.kr}

\author{Yunyong Ko}
\authornote{Corresponding author}% \authornotemark[1]
\affiliation{%
  \institution{Chung-Ang University}
  \city{Seoul}
  \country{Korea}}
\email{yyko@cau.ac.kr}
\orcid{0000-0003-1283-4697}


\begin{abstract}
  Knowledge graph completion (KGC) models are commonly evaluated using rank-based metrics such as MRR and Hits@K, 
  despite different users often requiring different evaluation perspectives. 
  In this demo, we present \textbf{{\method}}, an interactive system for probing diverse evaluation landscapes for KGC models. 
  {\method} enables users to flexibly evaluate KGC models by adjusting two critical perspectives: \textbf{(P1)} \textit{predictive sharpness} and \textbf{(P2)} \textit{popularity-bias robustness}. 
  Through a user-friendly GUI, users easily evaluate multiple KGC models and analyze their strengths and weaknesses.
  {\method} provides four key functionalities: (1) conventional evaluation toolkit, (2) flexible perspective-aware evaluation, (3) explainable case studies, and (4) evaluation landscape exploration. 
  We believe that {\method} can help users better understand KGC models aligning with their objectives.
\end{abstract}






\maketitle

\section{Introduction}\label{sec:intro}
Knowledge graph completion (KGC) plays a crucial role in improving the quality of KG-driven applications, 
including question answering~\cite{huang2019kgqa-www,yasunaga2021kgqa-acl}, recommender systems~\cite{wang2018rec-www,zhou2020rec-sigir,wang2019rec-www}, and drug discovery~\cite{zhang2021drug,lin2020drug-ijcai}.
Since the quality of predicted facts directly affects downstream applications, 
\textit{the evaluation of KGC models} is crucial for selecting appropriate models in practice.

A key challenge in KGC evaluation is that the notion of a ``good" KGC model often varies across difference users and applications. 
For example, practitioners in drug discovery may prefer models that can make highly accurate predictions with strong confidence as inaccurate predictions can lead to costly medical trials~\cite{zeng2022toward,vella2022medtrialcost}.
On the other hand, KGC researchers may be more interested in discovering novel knowledge related to rare entities and relations, rather than already well-known facts~\cite{zhou2020rec-sigir}.
As a result, different users may favor different KGC models depending on their evaluation perspectives.

However, existing rank-based metrics such as MRR and Hits@K evaluate KGC models from a single fixed perspective~\cite{bordes2013translating,sun2019rotate,zhang2022redgnn}, 
making it difficult to identify the most suitable model for a target application. 
As a result, users may select suboptimal KGC models.

\begin{figure}[t]
    \centering
    \includegraphics[width=0.95\linewidth]{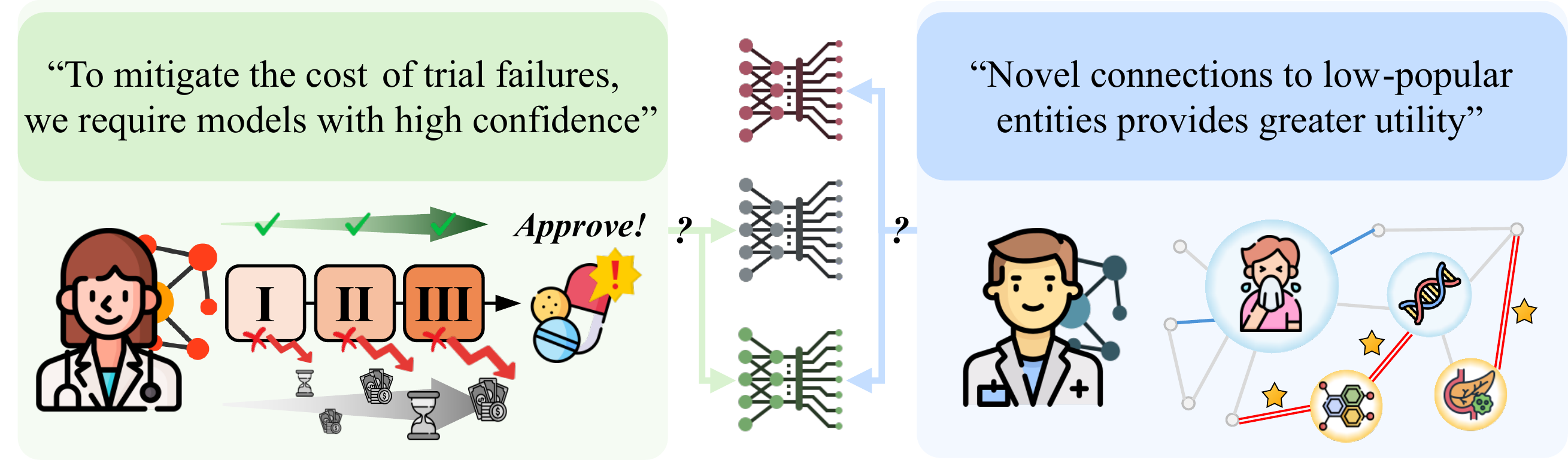}
    \vspace{-3mm}
    \caption{Motivation: different users may favor different KGC models depending on their evaluation perspectives.}
    \vspace{-5mm}
    \label{fig:fig1}
\end{figure}

To bridge the gap between diverse evaluation perspectives and existing fixed evaluation protocols, 
we develop \textbf{{\method}}, an interactive web-based system for probing evaluation landscapes of KGC models. 
Through a user-friendly GUI, {\method} enables users to evaluate KGC models from diverse perspectives beyond conventional metrics such as MRR and Hits@K, while providing insights into the relative strengths and weaknesses of KGC models.

{\method} is built on top of \textsf{PROBE}~\cite{moon2025sharp}, a generalized KGC evaluation framework that incorporates two key evaluation perspectives: \textbf{(P1)} \textit{predictive sharpness} and \textbf{(P2)} \textit{popularity-bias robustness}. 
Predictive sharpness controls how strongly inaccurate predictions are penalized, 
whereas popularity-bias robustness reflects the importance assigned to predictions involving low-popularity entities. 
By adjusting these two perspectives, 
{\method} enables users to probe diverse evaluation landscapes for KGC models.

\vspace{1mm}
\noindent
\textbf{System features of {\method}}.
{\method} provides \textbf{(1)} a user-friendly GUI that enables users easily to evaluate KGC models without additional programming effort;
\textbf{(2)} high accessibility, allowing users to access the system from anywhere on the web;
and \textbf{(3)} high portability, as it is independent of the underlying KGC model architectures, training frameworks, and execution environments, provided that users supply prediction results in our predefined input format (see Section~\ref{sec:demo} for details).

\vspace{1mm}
\noindent
\textbf{Evaluation functionalities of {\method}}.
{\method} provides the following four key functionalities for KGC evaluation:
\begin{itemize}[leftmargin=10pt]
    \item \textbf{(1) Conventional evaluation toolkit}: consisting of conventional rank-based metrics, six popular KGs, and pre-computed results of six state-of-the-art KGC models.
    \item \textbf{(2) Flexible perspective-aware evaluation}: allowing users to interactively adjust (P1) predictive sharpness and (P2) popularity-bias robustness according to user requirements.
    \item \textbf{(3) Explainable evaluation via case studies}: providing representative case studies that explain why a model performs well or poorly under a specific evaluation perspective
    \item \textbf{(4) Evaluation landscape exploration}: offering comprehensive visualization of evaluation landscapes across different levels of predictive sharpness and popularity-bias robustness, enabling users to probe model behaviors under diverse real-world application requirements.
\end{itemize}
Through these functionalities, {\method} helps users understand not only which KGC model performs best, but also why and under what evaluation perspectives it should be selected.

The source code and demo video of {\method} are available at:
\begin{itemize}[leftmargin=15pt]
    \item Code: \codeurl
    \item Video: \videourl
\end{itemize}

\section{Preliminary: \textsf{PROBE}}\label{sec:preliminary}

\noindent
\textbf{Overview}.
\textsf{PROBE} follows the standard rank-based evaluation protocol~\cite{sun2020reevaluation,qu2020rnnlogic,li2022house,bordes2013translating}.
Given a trained model $\theta(\cdot)$  and a set of test triples $\mathcal{T}_{test}$, 
\textsf{PROBE} consists of (1) prediction, (2) transformation, and (3) aggregation.

\begin{itemize}[leftmargin=10pt]
    \item \textbf{(1) Prediction}: For each query $(h,r,t)  \in \mathcal{T}_{test}$, the model $\theta(\cdot)$ ranks all candidate entities and produces the rank of the correct entity. A set of ranks $\mathbf{r}={r_1,r_2,\dots,r_n}$ is obtained.
    \item \textbf{(2) Transformation}: A rank transformation function $f:\mathbb{N}^1 \rightarrow \mathbb{R}^1$ converts each rank $r_i \in \mathbf{r}$ into a score $c_i=f(r_i)$. This step produces a set of transformed scores $\mathbf{c}={c_1,c_2,\dots,c_n}$, where better ranks generally receive higher scores.
    \item \textbf{(3) Aggregation}: An aggregation function $agg:\mathbb{R}^{n}\rightarrow\mathbb{R}^{1}$ computes the final evaluation score by taking the weighted average of all scores in $\mathbf{c}$.
\end{itemize}

\begin{figure}[t]
\centering
\setlength\tabcolsep{0pt}
\begin{tabular}{cc}
    \includegraphics[width=0.425\linewidth]{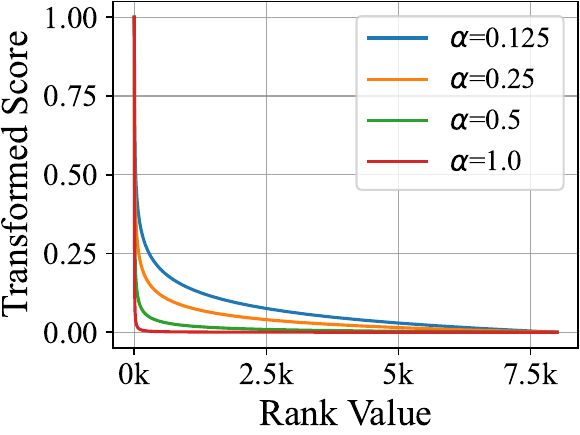} \qquad\qquad 
    &
    \includegraphics[width=0.425\linewidth]{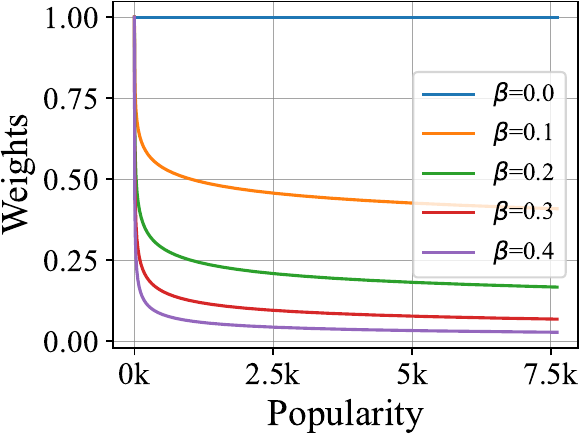} \\
    (a) RT function & (b) Weight function \\
\end{tabular}
\vspace{-3mm}
\caption{RT and RA functions of \textsf{PROBE}. Users can flexibly evaluate KGC models from diverse evaluation perspectives by adjusting $\alpha$ and $\beta$.}
\vspace{-5mm}
\label{fig:RT-RA}
\end{figure}

\vspace{1mm}
\noindent
\textbf{Rank transformation (RT)}.
The RT function converts each predicted rank $r \in \mathbf{r}$ into a score while controlling the desired level of predictive sharpness. 
Given a rank $r$ and a sharpness control factor $\alpha$, 
the RT function computes the transformed score as
\begin{equation}
    f^*(r,\alpha)=\frac{f(r,\alpha)-1}{1-(|\mathcal{E}|^{-\alpha})}+1\label{eq:transform-affine},
\end{equation}
where $|\mathcal{E}|$ denotes the number of entities in a KG. 
The parameter $\alpha$ controls how strongly inaccurate predictions are penalized. 
As shown in Figure~\ref{fig:RT-RA}(a),
% shows the transformed scores of the RT function according to the sharpness control factor $\alpha$.
larger values of $\alpha$ assign substantially lower scores to incorrect predictions, 
whereas smaller values produce a more gene evaluation.

\vspace{1mm}
\noindent
\textbf{Rank aggregation (RA)}.
Then, the RA function assigns different weights to predictions according to the popularity of their target entities. 
Intuitively, predictions associated with highly popular entities receive lower weights, 
as such entities are more likely to be sufficiently learned during training and can lead to popularity-biased evaluation.
Given a query $q$, its weight is defined as
\begin{align}
w_q = {1 \over (\epsilon + \delta(q)_{train})^{\beta}},
\end{align}
where $\delta(q)_{train}$ denotes the popularity of the target entity in the training set, $\beta$ controls the level of popularity-bias robustness, and $\epsilon$ is a small constant. 
Figure~\ref{fig:RT-RA}(b) shows the weight function according to the varying levels of popularity-bias robustness $\beta$.
Larger values of $\beta$ assign lower weights to high-popularity entities, thereby low-popularity entities contributing the final score more.
Finally, all scores $\mathbf{c}$ transformed in the RT are averaged with their weights $\mathbf{w}$ to compute the accuracy of a KGC model, which is defined as ${1 \over W}\sum^{|\mathcal{T}_{test}|*2}_{i=1} w_i \cdot c_i,\label{eq:aggregation}$.

% \begin{align}
%     {1 \over W}\sum^{|\mathcal{T}_{test}|*2}_{i=1} w_i \cdot c_i,\label{eq:aggregation}.
% \end{align}

\begin{figure*}[t]
    \centering
    \includegraphics[width=1.0\textwidth]{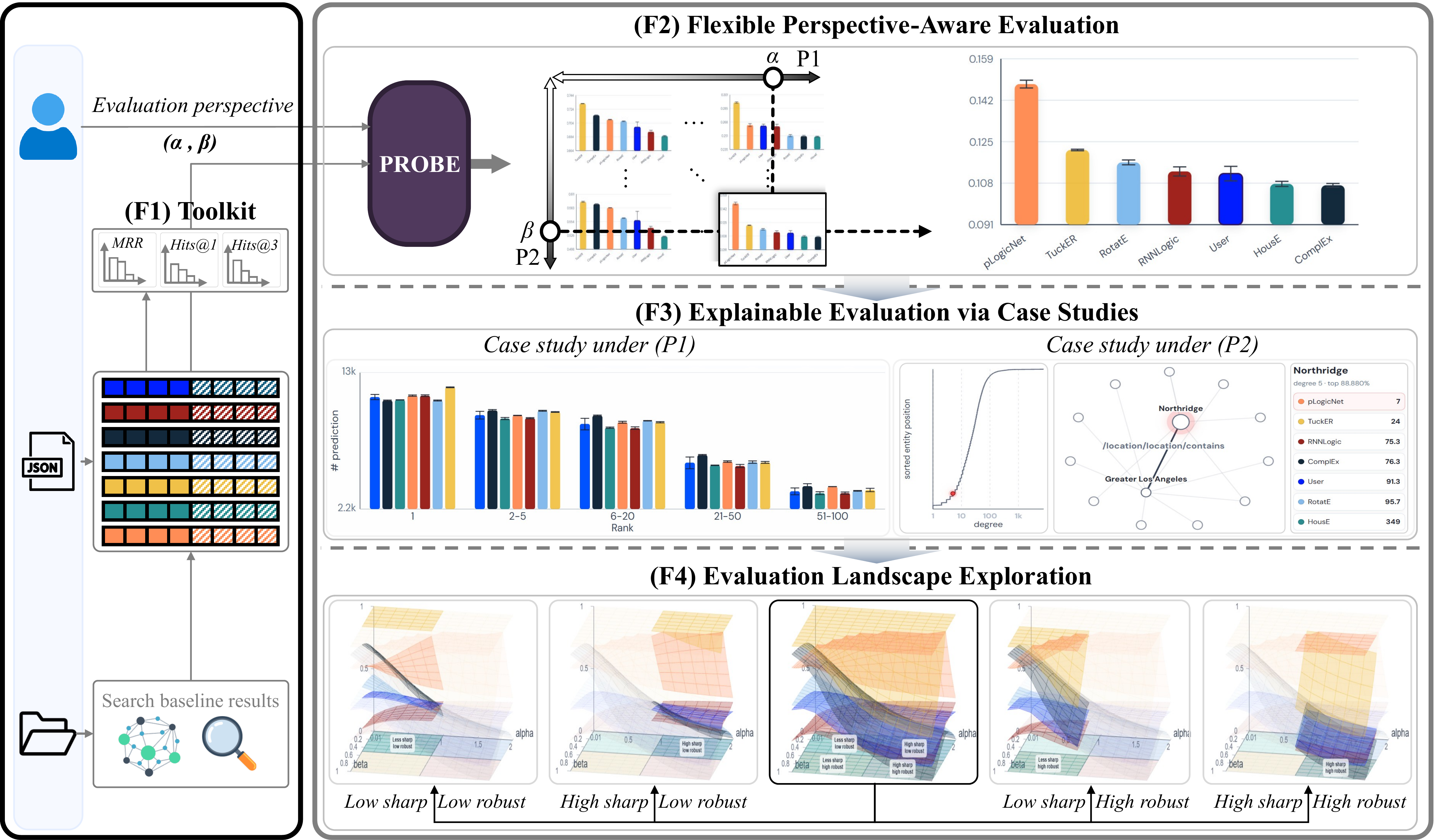}
    \vspace{-2mm}
    \caption{Overview of {\method} that provides the four key functionalities: (F1) Conventional Evaluation Toolkit, (F2) Flexible Perspective-Aware Evaluation, (F3) Explainable Evaluation via Case Studies, and (F4) Evaluation Landscape Exploration.}
    % \vspace{-2mm}
    \label{fig:overview}
\end{figure*}

\section{Demonstration of {\method}}\label{sec:demo}
\noindent
\textbf{Overview of {\method}}.
As illustrated in Figure~\ref{fig:overview},
{\method} consists of two stages: (1) Data Preparation and (2) Interactive Evaluation.
In the stage (1), 
users upload the prediction results of KGC models (i.e., rank scores) and the KG information (i.e., popularity statistics) in a predefined format. 
Specifically, the prediction results contain the ranked candidate entities generated by KGC models for each test query. 
Then, {\method} automatically analyze the prediction results of KGC models by using \textsf{PROBE}.

In the stage (2), 
{\method} provides a user-friendly GUI through which users can easily explore the four analysis functionalities: 
\textit{(1) Conventional Evaluation Toolkit}, \textit{(2) Flexible Perspective-Aware Evaluation}, \textit{(3) Explainable Evaluation via Case Studies}, and \textit{(4) Evaluation Landscape Exploration}. 
Through these functionalities, users can evaluate KGC models from diverse evaluation perspectives and gain deeper insights into their strengths and weaknesses.

\vspace{1mm}
\noindent
\textbf{Data Preparation}.
To evaluate KGC models in {\method}, users are required to upload four types of input files: 
(1) entity information, (2) relation information, (3) KG training triples, and (4) KGC model prediction results. 
These files are used to evaluate KGC models under diverse evaluation perspectives.

\begin{itemize}[leftmargin=10pt]
    \item \textbf{(1)-(3) KG information.} Users first upload three files describing the target KG: \texttt{entities.dict}, \texttt{relations.dict}, and \texttt{train.txt}. 
    The \texttt{entities.dict} and \texttt{relations.dict} files contain the ID and names of entities and relations, respectively:
    \begin{quote}
    \small
    \texttt{entities.dict}: \\
    \texttt{\{id\}\textbackslash t\{entity1\_name\}\textbackslash n} \texttt{\{id\}\textbackslash t\{entity2\_name\}\textbackslash n} $\cdots$ 
    
    \vspace{1mm}
    \texttt{relations.dict}: \\
    \texttt{\{id\}\textbackslash t\{relation1\_name\}\textbackslash n} \texttt{\{id\}\textbackslash t\{relation2\_name\}\textbackslash n} $\cdots$ 
    \end{quote}
    
    The \texttt{train.txt} file contains KG triples in the following format:
    \begin{quote}
    \small
    \texttt{\{head\_entity\_id\}\textbackslash t\{relation\_id\}\textbackslash t\{tail\_entity\_id\}\textbackslash n}$\cdots$ 
    \end{quote}  

    \item \textbf{(4) KGC model prediction results.} For each KGC model, users upload a JSON file containing prediction results on test queries. If multiple models are to be compared, multiple JSON files can be uploaded simultaneously. Each JSON file consists of a list of prediction records:
    \begin{quote}
    \small
    \texttt{[[test\_triple,\ mode,\ rank],\ ...,]}
    \end{quote}
    where \texttt{test\_triple = [h\_id,\ r\_id,\ t\_id]} denotes a test triple,
    \texttt{mode} is either \texttt{"h"} or \texttt{"t"}, indicating head-entity prediction or tail-entity prediction,
    and \texttt{rank} is the rank assigned to the correct entity by the KGC model ($rank \geq 1$).
\end{itemize}
Based on these files, {\method} computes the required statistics (i.e., popularity of entities) and enables all evaluation and visualization functionalities through its interactive interface.

\begin{table}[t]
\centering
\small
\caption{(F1) Built-in evaluation toolkit of {\method}.}
\vspace{-3mm}
\label{table:toolkit}
\setlength\tabcolsep{3pt}
\def\arraystretch{0.98} % row space
\begin{tabular}{c|c}
\toprule
\textbf{Metrics} & MRR, Hits@1, Hits@3 \\

\midrule
\multirow{2}{*}{\textbf{KG}} & FB15k-237~\cite{toutanova2015observed}, WN18RR~\cite{dettmers2018convolutional}, YAGO3-10~\cite{mahdisoltani2013yago3} \\
 & Family~\cite{yang2017differentiable}, UMLS~\cite{nickel2011rescal}, Kinship~\cite{nickel2011rescal} \\
 
\midrule
\textbf{Pre-computed results} & RotatE~\cite{sun2019rotate}, ComplEx~\cite{trouillon2016complex}, HousE~\cite{li2022house} \\
 \textbf{of KGC models} & TuckEr~\cite{balazevic2019tucker}, pLogicNet~\cite{qu2019probabilistic}, and RNNLogic~\cite{qu2020rnnlogic} \\
\bottomrule
\end{tabular}
\end{table}

\begin{figure}[t]
    \centering
    \includegraphics[width=0.96\linewidth]{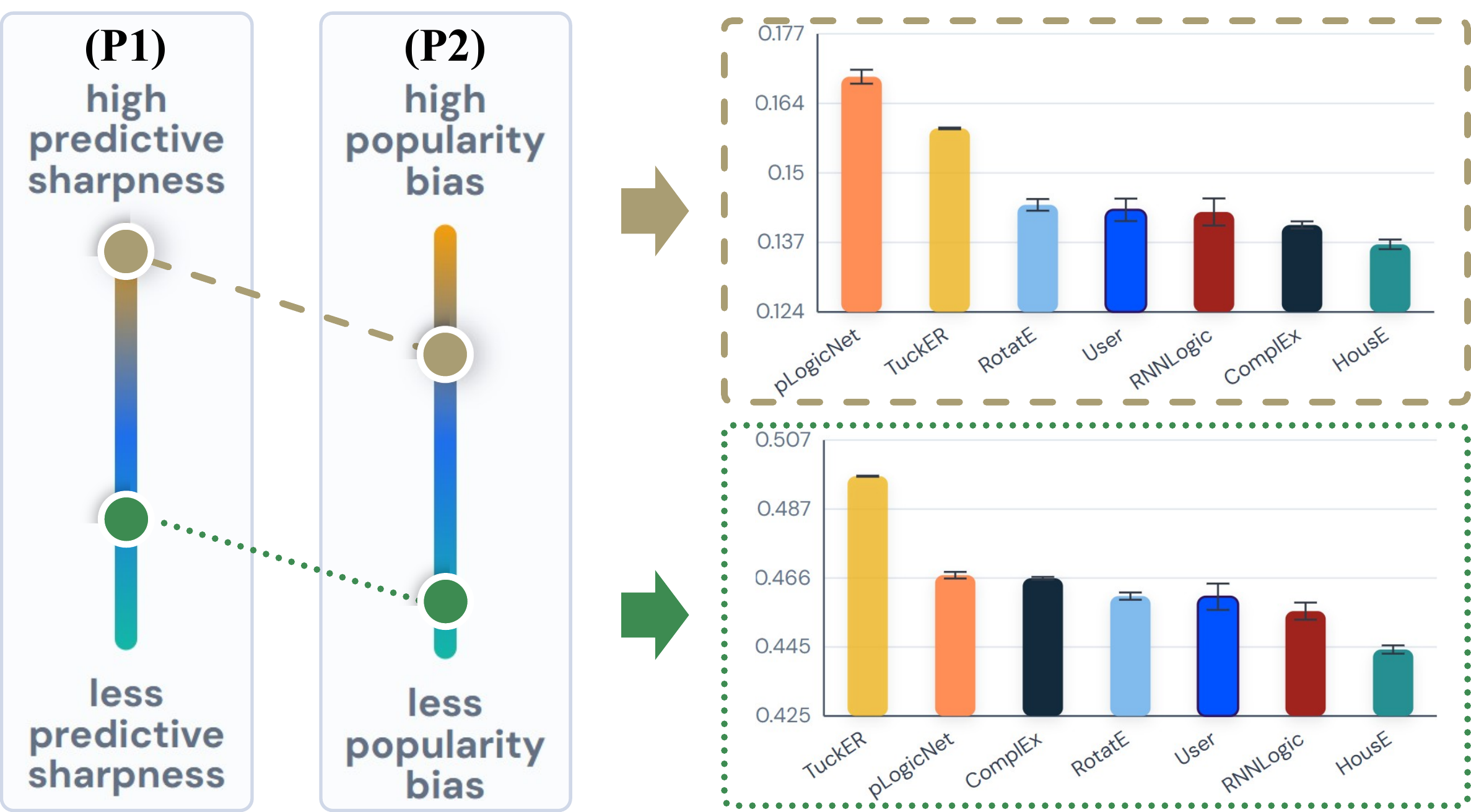}
    \vspace{-2mm}
    \caption{(F2) Flexible perspective-aware evaluation. By interactively adjusting $\alpha$ and $\beta$, users can observe how the relative rankings of models change under different perspectives.}
    \vspace{-3mm}
    \label{fig:flexible-evaluation}
\end{figure}

\vspace{1mm}
\noindent
\textbf{(F1) Conventional evaluation toolkit}.
The conventional evaluation toolkit provides a standard benchmark view of KGC model performance using widely adopted rank-based metrics, including MRR, and Hits@K. 
Given uploaded prediction results of KGC models, {\method} presents the relative ranking of KGC models under conventional evaluation protocols.
To facilitate immediate exploration without requiring user-provided data, 
{\method} also includes six widely used benchmark KGs together with pre-computed prediction results of six KGC models (see Table~\ref{table:toolkit}).
As a result, users can directly compare existing KGC models under conventional metrics and familiarize themselves with the evaluation interface even without uploading their own KGs or prediction results.

\begin{figure*}[t]
    \centering
    \includegraphics[width=0.98\linewidth]{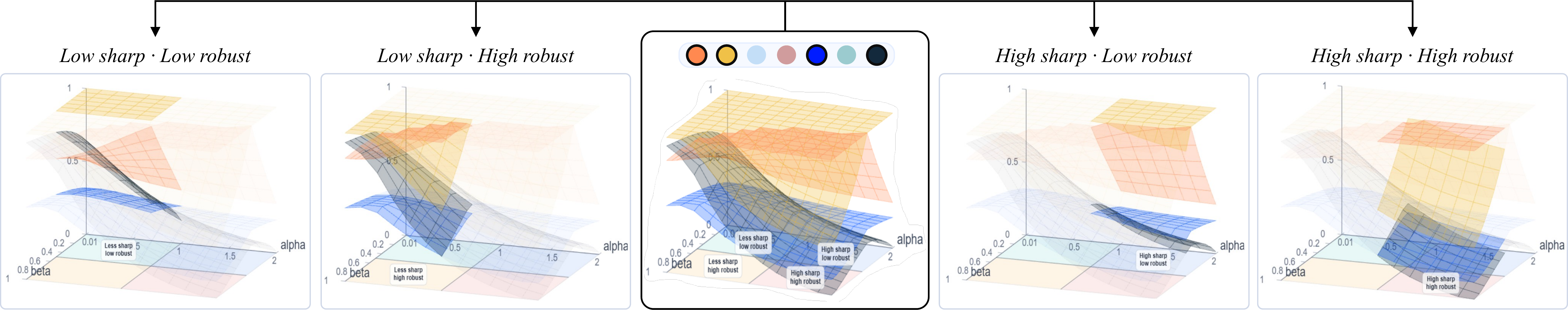}
    \caption{Evaluation landscape exploration. {\method} visualizes the performance of KGC models in a 3-D evaluation landscape defined by predictive sharpness ($\alpha$) and popularity-bias robustness ($\beta$). By comparing model surfaces, users can identify the relative strengths, weaknesses, and performance trends of KGC models under diverse evaluation perspectives at a glance.}
    \label{fig:landscape}
\end{figure*}

\vspace{1mm}
\noindent
\textbf{(F2) Flexible perspective-aware evaluation}.
As illustrated in Figure~\ref{fig:flexible-evaluation}, 
this functionality allows users to interactively adjust the predictive sharpness factor $\alpha$ and the popularity-bias robustness factor $\beta$ through the graphical interface. 
As users modify these factors, {\method} automatically recomputes evaluation scores and updates the relative rankings of KGC models in real time. 
This enables users to easily investigate how model rankings change under different evaluation perspectives and identify models that best align with their evaluation objectives.

\begin{figure}[t]
    \centering
    \includegraphics[width=0.98\linewidth]{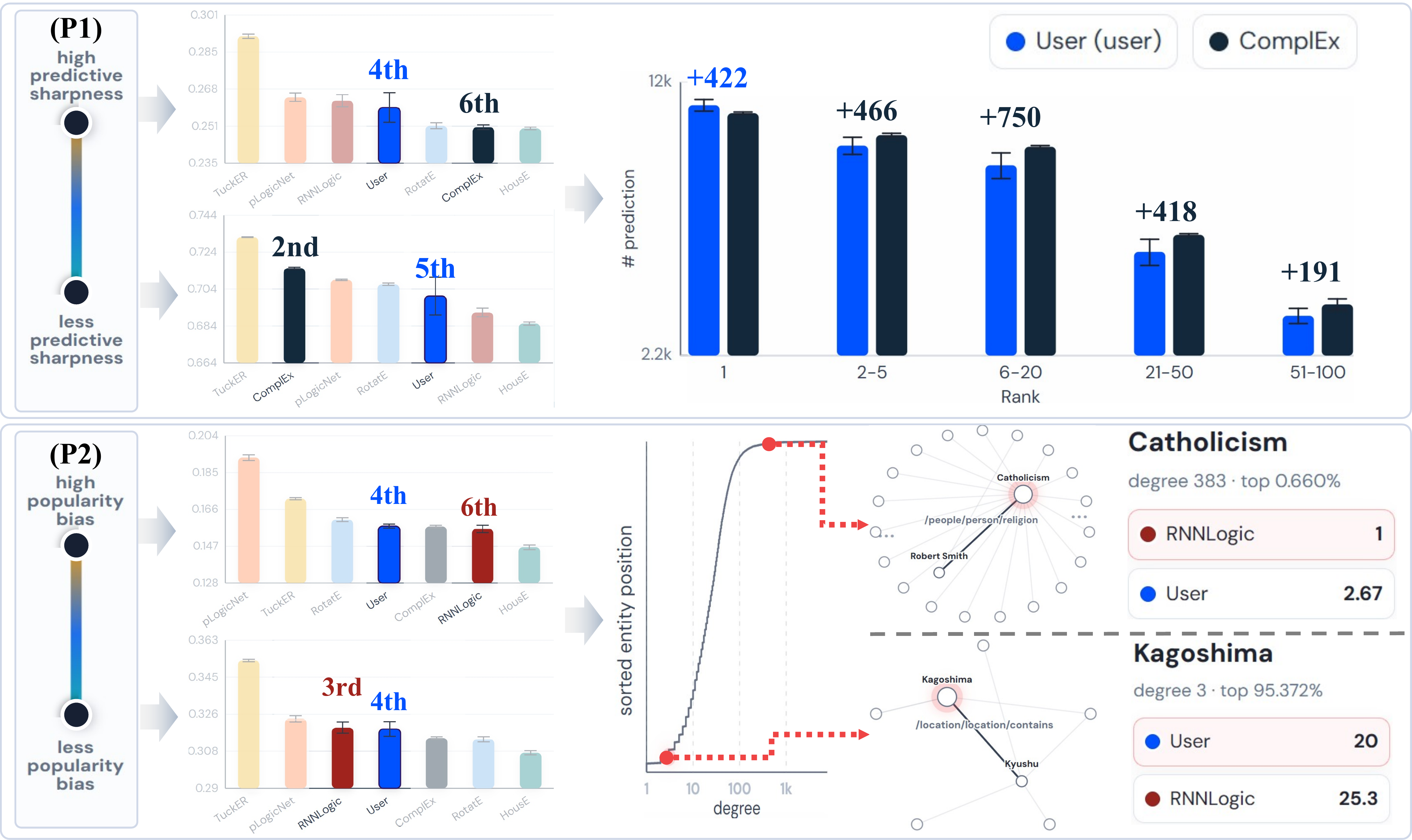}
    \vspace{-3mm}
    \caption{(F3) Explainable evaluation via case studies. {\method} explains ranking changes under different evaluation perspectives through rank-group analysis (P1) and popularity-aware query-level case studies (P2).}
    \label{fig:case-study}
    \vspace{-4mm}
\end{figure}

\vspace{1mm}
\noindent
\textbf{(F3) Explainable evaluation via case studies}.
Although (F2) enables users to observe changes in model rankings under different evaluation perspectives, 
understanding \emph{why} such ranking changes occur remains challenging. 
To address this issue, {\method} provides explainable evaluation via case studies from both the predictive sharpness (P1) and popularity-bias robustness (P2) perspectives.
As illustrated in Figure~\ref{fig:case-study}, the interface consists of two panels corresponding to the two evaluation perspectives. 

In the \textbf{P1 panel}, users can adjust the predictive sharpness factor $\alpha$ and observe the resulting changes in model rankings. 
To explain these changes, {\method} presents the prediction quality of each model across different rank groups (e.g., rank 1st, ranks 2nd-5th, and ranks 6th-20th), 
allowing users to identify whether a model produces top-rank predictions or consistently generates moderately accurate predictions.

Similarly, in the \textbf{P2 panel}, users can adjust the popularity-bias robustness factor $\beta$ and observe the corresponding ranking changes. 
The system further provides representative query-level case studies, 
including the popularity of individual queries and the ranks assigned by different KGC models. 
Through these examples, users can easily examine how models behave on low-popularity queries and understand their relative strengths and weaknesses under popularity-aware evaluation.

\vspace{1mm}
\noindent
\textbf{(F4) Evaluation landscape exploration}. 
Finally, {\method} provides a \textit{holistic} view of model performance across the entire \textit{evaluation landscape}. 
As illustrated in Figure~\ref{fig:landscape}, {\method} visualizes the performance of KGC models in a 3-D space, 
where the $x$-axis and $y$-axis represent the predictive sharpness factor $\alpha$ and the popularity-bias robustness factor $\beta$, respectively, and the $z$-axis represents the resulting normalized evaluation score.
Under this representation, each KGC model forms a performance surface in the evaluation landscape. 
When multiple KGC models are selected, their surfaces are displayed simultaneously, 
enabling direct comparison across diverse evaluation perspectives. 
As shown in Figure~\ref{fig:landscape}, one model may achieve higher performance in certain regions of the landscape, 
while another model may dominate in different regions. 

Furthermore, {\method} provides a detailed quadrant view of the evaluation landscape by partitioning the $(\alpha,\beta)$ space into four representative regions. 
This functionality allows users to closely inspect KGC model behaviors under representative evaluation perspectives and facilitates more fine-grained comparison.
Through this visualization, users can not only compare KGC models at a glance, but also understand their relative strengths, weaknesses, and performance trends across a wide range of evaluation perspectives.

\vspace{-1.5mm}
\section{Conclusion}\label{sec:conclusion}
In this demo, we present {\method}, an interactive web-based system for probing evaluation landscapes of KGC models.
Built on top of the \textsf{PROBE} framework, 
{\method} enables users to easily evaluate and analyze KGC models under diverse evaluation perspectives through a user-friendly GUI.
By providing conventional evaluation, perspective-aware analysis, explainable case studies, and evaluation landscape exploration, 
{\method} helps users better understand the strengths and weaknesses of KGC models and select models that best match their evaluation objectives.

\vspace{-1.5mm}
\begin{acks}
This work was supported by the National Research Foundation of Korea (NRF) grant funded by the Korea government (MSIT) (RS-2024-00459301).
\end{acks}

% \clearpage

\section*{GenAI Usage Disclosure}
In accordance with the ACM authorship policy, we disclose the usage of generative AI tools (e.g., ChatGPT) as follows.
\begin{itemize}[leftmargin=10pt]
    \item \textbf{GenAI usage in writing:} ChatGPT was used exclusively to review grammatical consistency during the writing of the manuscript.
    \item \textbf{GenAI usage in data processing:} ChatGPT was only used for generating code for drawing figures (e.g., matplotlib.pyplot).
    \item \textbf{GenAI usage in programming:} ChatGPT was only used for assisting with code implementation for experiments.
    \item \textbf{Author responsibility:} All uses of GenAI were limited to assistant roles. We conducted final verification and refinement of all GenAI generated results, analyses, and textual content.
\end{itemize}

\bibliographystyle{ACM-Reference-Format}
\balance
\bibliography{bibliography.bib}

\end{document}